\documentclass[conference]{IEEEtran}
\IEEEoverridecommandlockouts

\usepackage{cite}
\usepackage{amsmath,amssymb,amsfonts}
\usepackage{algorithmic}
\usepackage{graphicx}
\usepackage{textcomp}
\usepackage{xcolor}
\usepackage{booktabs}
\usepackage{multirow}
\usepackage{float}
\usepackage{tabularx}
\usepackage{array}
\usepackage{adjustbox}
\usepackage{ltablex} 
\usepackage{longtable} 
\usepackage{tabularx} 
\usepackage{hyperref}
\usepackage{adjustbox}
\usepackage{lipsum} 
\usepackage{tabto}
\usepackage{xcolor,colortbl}
\usepackage{tabularx}
\newcommand{\framework}{\textsc{ClinicSum}\xspace}
\usepackage{pifont}
\usepackage[utf8]{inputenc}
\usepackage{array}
\usepackage{soul}
\usepackage{comment}
\usepackage[most]{tcolorbox}
\usepackage[T1]{fontenc}
\usepackage{caption}
\usepackage{subcaption}
\usepackage{xspace}
\usepackage{tikz}
\definecolor{color1}{HTML}{34FF34}
\definecolor{color2}{HTML}{67FD9A}
\definecolor{color3}{HTML}{FD6864}

\def\BibTeX{{\rm B\kern-.05em{\sc i\kern-.025em b}\kern-.08em
    T\kern-.1667em\lower.7ex\hbox{E}\kern-.125emX}}
\begin{document}

\title{\framework: Utilizing Language Models for Generating Clinical Summaries from Patient-Doctor Conversations}


\author{\IEEEauthorblockN{Subash Neupane\IEEEauthorrefmark{1},  Himanshu Tripathi\IEEEauthorrefmark{2},
Shaswata Mitra\IEEEauthorrefmark{3},
Sean Bozorgzad\IEEEauthorrefmark{4}, \\
Sudip Mittal\IEEEauthorrefmark{5},
Shahram Rahimi\IEEEauthorrefmark{6}, and
Amin Amirlatifi\IEEEauthorrefmark{7}
}

\IEEEauthorblockA{Dept. of Computer Science and Engineering,
Mississippi State University\\
Potentia Analytics Inc.\\
Dave C. Swalm School of Chemical Engineering,
Mississippi State University\\
Email: \{\IEEEauthorrefmark{1}sn922,
\IEEEauthorrefmark{2}ht577,
\IEEEauthorrefmark{3}sm3843\}@msstate.edu,
\{\IEEEauthorrefmark{4}sean\}@potentiaco.com}
\{\IEEEauthorrefmark{5}mittal,
\IEEEauthorrefmark{6}rahimi\}@cse.msstate.edu,
\IEEEauthorrefmark{7}amin@che.msstate.edu
}

\maketitle

\begin{abstract}

This paper presents \framework, a novel framework designed to automatically generate clinical summaries from patient-doctor conversations. It utilizes a two-module architecture: a retrieval-based filtering module that extracts Subjective, Objective, Assessment, and Plan (SOAP) information from conversation transcripts, and an inference module powered by fine-tuned Pre-trained Language Models (PLMs), which leverage the extracted SOAP data to generate abstracted clinical summaries. To fine-tune the PLM, we created a training dataset of consisting 1,473 conversations-summaries pair by consolidating two publicly available datasets, FigShare and MTS-Dialog, with ground truth summaries validated by Subject Matter Experts (SMEs). \framework's effectiveness is evaluated through both automatic metrics (e.g., ROUGE, BERTScore) and expert human assessments. Results show that \framework outperforms state-of-the-art PLMs, demonstrating superior precision, recall, and F-1 scores in automatic evaluations and receiving high preference from SMEs in human assessment, making it a robust solution for automated clinical summarization.


\end{abstract}


\section{Introduction}
\label{introduction}

The advent of transformer-based models such as OpenAI GPT models \footnote{https://platform.openai.com/docs/models}, Meta LLAMA\footnote{https://llama.meta.com/} variants, and Google Gemini\footnote{https://gemini.google.com} has revolutionized Natural Language Processing (NLP) by significantly improving performance across a wide array of tasks. These advancements, driven primarily by transfer learning, have opened up new possibilities for applying these models in specialized domains\cite{neupane2024medinsight, zhao2024retrieval}. One such domain is healthcare, where leveraging Pre-trained Language Models (PLMs) to automatically generate clinical summaries from doctor-patient conversations presents a promising application with substantial benefits for both patients and healthcare providers.

Clinical summaries play a critical role in healthcare by improving patients' understanding of care plans and reducing the risk of misinterpreting medical information. Research indicates that patients forget 40-80\% of the medical information provided by healthcare practitioners almost immediately \cite{mcguire1996remembering} and misconstrue nearly half of what they remember \cite{anderson1979patient}. For healthcare providers, generating these summaries automatically can alleviate the administrative burden of updating Electronic Health Records (EHRs), a task strongly associated with physician burnout \cite{kumar2016burnout, knoll2022user}.

\begin{figure}
  \centering
  \includegraphics[width=1.0\linewidth]{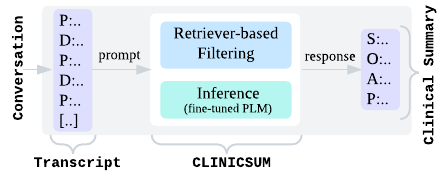}
  \caption{A graphical overview of the \framework. \emph{P} denotes the Patient and \emph{D} denotes the Doctor in the conversation transcript. \emph{S, O, A,} and \emph{P} refer to the Subjective, Objective, Assessment, and Plan components of the clinical summary.}
  \label{fig:overview_dialogsum}
  \vspace{-5mm}
\end{figure}

However, the application of PLMs in this context is not without challenges. Since PLMs are generally trained on broad, non-specialized text corpora, they are prone to producing inaccuracies—such as hallucinations\cite{huang2023survey}—that could have serious consequences for patient care. Addressing these challenges requires more than just deploying PLMs; it necessitates a tailored approach that can accurately capture the nuances of medical conversations while ensuring the reliability of the generated summaries.

In this paper, we present \framework, a comprehensive framework designed to automatically generate clinical summaries in the Subjective Objective Assessment and Plan (SOAP) format from transcribed patient-doctor conversations. Fig. \ref{fig:overview_dialogsum} provides an illustration of \framework. To tackle the limitations of current PLMs in healthcare, \framework integrates a retrieval-based filtering module and an inference module, working in tandem to produce accurate and contextually relevant summaries. The retrieval-based filtering module is responsible for extracting the SOAP components from the given transcript which it achieves by leveraging an ensemble retriever approach, that combines sparse and dense retrieval techniques, to capture both lexical and semantic meanings from the transcribed conversations. By utilizing this dual approach, filtering method ensures that the most relevant information is passed to the inference module, which is fine-tuned to generate clinical summaries.

For fine-tuning a PLM, we collaborated with Potentia Inc.\footnote{https://www.potentiaco.com/}, a healthcare software company, to create a new training dataset. This dataset was constructed by combining 1,473 patient-doctor conversations from two publicly available sources, FigShare\footnote{figsahre.com} and MTS-Dialog\cite{mts-dialog} and generating their 
corresponding clinical summaries. Subject Matter Experts (SMEs), including doctors and physicians from Potentia Inc., manually reviewed and corrected the summaries to ensure their high quality, providing reliable data for fine-tuning task. The final training data is publicly available through huggingface \footnote{huggingface.co/datasets/SubashNeupane/dataset\_SOAP\_summary}.

Previous research on generating clinical summaries includes Zhang et al.\cite{zhang2021leveraging}, which fine-tuned a BART model to handle long and noisy doctor-patient conversation transcripts, and Giorgi et al.\cite{giorgi2023wanglab}, which used fine-tuning and few-shot In-Context Learning (ICL)\cite{brown2020language} with GPT-4. While these methods have achieved some success, \framework introduces a unique combination of retrieval-based filtration and a fine-tuned inference module to generate SOAP format clinical summaries. Unlike the multistage summarization process for long conversations used by Zhang et al. or the focus on ICL by Giorgi et al., our approach refines input data before summarization, leading to superior performance in both automatic and expert human evaluations.

The main contributions of this paper are as follows:

\begin{itemize}

  \item We demonstrate the feasibility of generating clinical summaries in SOAP format utilizing the transcribed patient-doctor conversations. 
    \item We create a new dataset of clinical summaries corresponding to patient-doctor conversations from the FigShare and MTS-Dialog datasets.
        \item We built \framework - a framework that automatically generates clinical summaries. 
         \item We showcase \framework's proficiency in generating accurate and relevant clinical summaries through both automatic  and expert human evaluations. 
         \end{itemize}

The rest of the paper is organized as follows: Section \ref{background_related_work} discusses the background and related works. Section \ref{task_formulation} describes our task. Section \ref{method} provides insight on \framework's architecture and methodology. In Section \ref{experiment} we present our experiments, evaluation, and results. Section \ref{limitation} discusses the limitations and Section \ref{conclusion} concludes the paper.  

\section{Background \& Related Work}
\label{background_related_work}
\subsection{Clinical Summaries}

Clinical summaries are concise records of patient encounters, detailing medical history, current condition, treatment plans, and progress. Automatically generating these summaries from doctor-patient conversations benefits patients by improving recall and understanding \cite{giorgi2023wanglab} of care plans, and helps doctors by streamlining documentation \cite{schloss2020towards} and reducing administrative workload\cite{sinsky2016allocation}. One widely used format for clinical summary is SOAP\cite{weed1971problem}. 
An example of a clinical summary in SOAP format is presented in Fig. \ref{fig:soap_summaries}(B). The \emph{Subjective} section documents the patient's personal experiences, including the Chief Complaint (CC), History of Present Illness (HPI), and relevant medical history. The \emph{Objective} section records measurable data like vital signs, physical exam findings, and lab results. The \emph{Assessment} section combines subjective and objective information to diagnose the patient's condition, highlighting the problem and differential diagnoses. Lastly, the \emph{Plan} section details the approach for addressing or investigating the problem further. These structured summaries aids in making informed clinical decisions, tracking patient progress, and maintaining continuity of care \cite{neupane2024medinsight}.

\subsection{LLM, RAG and Fine-tuning}
Transformer architectures \cite{vaswani2017attention} have fueled the advancement of Large Language Models (LLMs) in NLP, thanks to their remarkable parallelization capabilities \cite{min2023recent}. Trained on massive internet text datasets and featuring substantial parameter sizes, LLMs exhibit impressive learning abilities. However, LLMs often struggle with factual questions in closed domains, where specialized knowledge is crucial. This difficulty can manifest in factually inaccurate predictions, a phenomenon known as \emph{hallucination}\cite{huang2023survey}. This limitation may arise from a combination of factors, including a \emph{deficit in domain knowledge}, \emph{reliance on outdated information}, and \emph{forgetting} \cite{wang2023survey, kirkpatrick2017overcoming}. 

To mitigate the knowledge deficiency within PLMs for domain-specific tasks, an additional knowledge ingestion step is required. The two most common approaches currently practiced for external knowledge ingestion are Retrieval Augmented Generation (RAG) and Fine-tuning. The first approach, introduced around mid-2020 by Lewis et al., \cite{lewis2020retrieval}, is designed to enhance the performance of PLMs on knowledge-intensive tasks. This approach involves retrieving relevant information from external knowledge sources based on the input query. The retrieved content is then concatenated with the original query, providing the PLM with enriched context, which leads to more informed and accurate response generation.

The second approach, is to fine-tune PLM. In this approach, a PLM is further trained on a smaller, task-specific dataset to adapt it to a particular application. This process allows the model to leverage the general knowledge it has acquired during pre-training and refine its weights based on the new, more focused data, improving its performance on the target task. As PLM grows in size, updating all parameters during fine-tuning becomes increasingly costly and inefficient, especially with limited computational resources. This challenge has driven research into Parameter Efficient Fine-Tuning (PEFT) methods that minimize tunable parameters while maintaining performance. Key approaches include adapter-based methods \cite{houlsby2019parameter}, prompt-based techniques \cite{lester2021power}, LoRA \cite{valipour2022dylora}, QLoRA \cite{dettmers2024qlora}.

In contrast to RAG and fine-tuning approaches, an alternative approach is ICL. This technique utilizes examples (usually few-shot) embedded within the prompt to guide the model's response generation.

\subsection{Related Works}

The field of open-domain dialogue summarization, encompassing the task of summarizing conversations and meetings, remains relatively unexplored. While there have been a limited number of studies dedicated to this area \cite{goo2018abstractive, li2019keep}, research interest in summarizing dialogue within closed domains, particularly in the medical field, has been gaining momentum in recent years. Specifically, the automatic generation of clinical summaries from doctor-patient conversations has attracted significant attention \cite{zhang2018learning, finley2018automated, schloss2020towards, enarvi2020generating, knoll2022user}.

To this day, various methods have been proposed, to generate clinical summaries in SOAP format including extractive and abstractive approaches. For example, Krishna et al. \cite{krishna2020generating} proposed a modular approach combining extractive and abstractive summarization techniques to generate SOAP notes from doctor-patient conversations. Building upon the work of \cite{krishna2020generating}, Ramprasad et al.,\cite{ramprasad2023generating} on the other hand, focused on enhancing the faithfulness and consistency of SOAP notes generated by LLMs. Their work introduces section-specific cross-attention parameters in encoder-decoder models to improve the factual accuracy and relevance of generated notes. While  Schloss and Konam \cite{schloss2020towards} concentrated on classifying utterances from medical conversations into SOAP sections and speaker roles using a hierarchical encoder-decoder model. 

In addition to extractive-abstractive methods, current research in automatic SOAP summary generation often involves fine-tuning PLM which closely aligns with our approach as discussed in Section \ref{method}. For example, Zhang et al., \cite{zhang2021leveraging} fine-tuned a pre-trained BART model to automatically generate summaries from doctor-patient conversations. However, their work was limited to just two specialties, internal medicine and primary care, and the training data included only the History of Present Illness (HPI) section. Similarly, Giorgi et al., \cite{giorgi2023wanglab} explored two approaches first fine-tuning a PLM (Longformer-Encoder-Decoder\footnote{https://huggingface.co/docs/transformers/en/model\_doc/led})  and second using few-shot ICL\cite{radford2019language}. In contrast, our approach combines retrieval-based filtering with inference using fine-tuned models in a zero-shot setting.

\section{Task Formulation}
\label{task_formulation}
Given a set of patient-doctor conversation transcripts \( T \), where each transcript \( t_i \in T \) consists of unstructured conversation. The objective is to generate semi-structured SOAP clinical summaries for each transcript $t_i$. This involves using a function \( f \), which applies a combination of information retrieval techniques and a PLM to map each \( t_i \) to a semi-structured SOAP format \( n_i \). Specifically, the function can be defined as:

\begin{equation}
    n_i = f(t_i) \forall t_i \in T, n_i \in N | i \in \mathbb{N}   
\end{equation}

\noindent where, $T$ is the set of Transcripts and $N$ is the set of clinical summaries. These summaries $n_i$ are organized in SOAP format where \( S \) is the \textbf{Subjective} component that summarizes patient's reported symptoms and experiences. \( O \) is the \textbf{Objective} component, detailing the observable and measurable clinical findings. \( A \) is the \textbf{Assessment} component, providing a diagnosis or evaluation based on the information and  \( P \) is the \textbf{Plan} component, outlining the treatment and management strategies discussed in \( t_i \).

\section{Architecture \& Methodology}
\label{method}

This section presents the architecture of our framework, \framework, and outlines our methodology. The framework consists of two main modules: \emph{retriever-based filtering} and \emph{inference}, as illustrated in Fig. \ref{fig:soap_summaries}. The following subsections provide a detailed explanation of each module.
\subsection{Retriever-based Filtering}
The first module in our framework systematically processes doctor-patient conversation transcripts $t_i$ to extract the SOAP elements for clinical summary $n_i$ using a retrieval prompt (query) $Q_R$. For example, our retrieval prompt is
\emph{``Extract subjective, objective, assessment, and plan details from a given transcript''}. 
For extraction this module utilizes the following three sub-components:\\


\subsubsection{\textbf{Splitting}} Splitting is the process of converting entire transcript $t_i$ into a set of individual sentences/chunks $c_i$.
To split $t_i$ into $c_i$, we apply a sentence split regular expression. 
Hence, the splitting ($t_i \rightarrow c_i$) can be formulated as:
\begin{equation}
    \{c_i : c_i \in C \}  = split(t_j) \forall t_j \in T | i,j \in \mathbb{N}, i \geq j
\end{equation}









\subsubsection{\textbf{Indexing}} Given the set of sentences ($C$) in $t_i$, where $c_i$ represents a sentence in the transcript, indexing is the process of projecting $c_i$ into vector space ($E$) through an embedding model $\xi(.)$, where, $e_i$ is the vector embedding of $c_i$ and we store this obtained vector embedding into vector storage.

\begin{equation}
    \{ e_i: e_i \in E \} = \xi(c_i) \forall c_i \in C | i \in \mathbb{N}
\end{equation}

\begin{figure*}
  \includegraphics[width=1\linewidth]{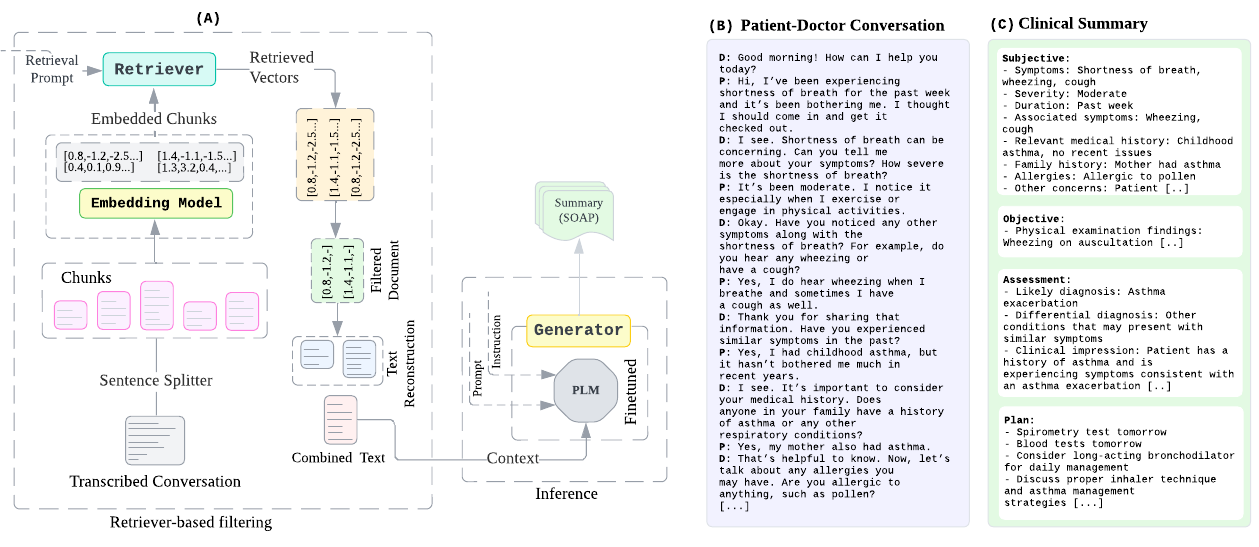}
  \caption{\textbf{A} is graphical illustration of the \framework architecture. It comprises two modules: \emph{retrieved-based filtering} and \emph{inference}. \textbf{B} represents patient-doctor conversation, and \textbf{C} represents generated clinical summary. [...] (used for brevity) indicates that there is more textual information.}
  \label{fig:soap_summaries}
  \vspace{-3mm}
\end{figure*}






\subsubsection{\textbf{Retrieval}} The retrieval process uses an ensemble method that combines sparse retriever (for example BM25 \cite{robertson1995okapi}) and dense retriever (for example DPR \cite{karpukhin2020dense} or our previous work \cite{tripathi2023experimental}), assigning different weights to each ($W_{Sparse}$, and $W_{Dense}$), then ranks it using ranking algorithm. In our use case, we implemented Reciprocal Rank Fusion (RRF) \cite{cormack2009reciprocal} which combines rankings from multiple sources by computing reciprocal rank scores. \\

A sparse retriever searches for documents ($c_i$) similar to $Q_R$ based on exact token matches in the sparse vector space ($C$), usually employing traditional keyword-based methods or indexing techniques. We employ BM25 as our sparse retriever ($R_{Sparse}$) that can be represented as:


\begin{equation}
    \{ c_{i\_Sparse} : c_{i\_Sparse} \in C | 1 \leq i \leq k \} = R_{Sparse}(C, Q_R)
\end{equation}
where, $k$ is the number of chunks with highest term frequency.\\



A dense retriever searches for documents relevant to $Q_R$ based on the exact or approximate neighbor similarity of embedded vectors ($e_i$) in a continuous embedding vector space ($E$), using dense representations. For retrieval, we also embed the retrieval prompt $Q_R$ such that:

\begin{equation}
    e_{Q_R} = \xi(Q_R)
\end{equation} 

\noindent Dense retriever ($R_{Dense}$) can be represented as:

\begin{equation}
  \{ e_i : e_i \in E, 1 \leq k \} = R_{Dense}(E, e_{Q_R})
\end{equation} 
The similarity function ($Sim (.)$) can be cosine, dot-product, or euclidean. The top $k$ relevant embedding ($e_i$) are then decoded ($c_i \leftarrow e_i$) to corresponding sentences ($c_i$). where the similarity function in $R_{Dense}$ is as follows:

\begin{equation}
Sim(e{_i}, Q_R) = \quad \frac{e_{Q_{R}} \cdot e_{c_i}}{\left\|e_{Q_{R}}\right\| \cdot\left\|e_{c_i}\right\|}
\end{equation}

In order to obtain the corresponding sentence/chunk ($c_i$) from embedding $e_i$, we apply an inverse embedding or decoding function ($\overline{\xi(\cdot)}$).

\begin{equation}
  \{ c_{i\_Dense} : c_{i\_Dense} \in C | 1\leq i \leq k \} = \{ \overline{\xi(e_i)}  \forall e_i \in E \}
\end{equation}









Next, we combine both $c_{i\_Sparse}$ and $c_{i\_Dense}$ to obtain the final set of embedded documents, before re-ranking them:
\begin{equation}
    c_{i\_Retrieved} = c_{i\_Dense} \cup c_{i\_Sparse}
\end{equation}

The cardinality of $c_{i\_Retrieved}$ (say $p$) will be less or equal to the total number of chunks retrieved using sparse and dense retriever ($k + k = 2k \ | \ k \in \mathbb{N}$) i.e., $p \leq |c_{i\_Sparse}| + |c_{i\_Dense}| \leq \mathbb{N}$. Once combined, we apply a ranking method to reorder the documents. This is done by using RRF algorithm. The algorithm works by calculating rank score ($r_i$) for the corresponding retrieved chunk ($c_{i\_Retrieved}$). If a document appears in both $c_{i\_Sparse}$ and $c_{i\_Dense}$ with different rankings, we sum the reciprocals of each rank from both retrievers. Typically $S(c_{i\_Sparse}), S(c_{i\_Dense}) \in [0, 1]$, this summed reciprocal score can exceed 1. This combined score is used for final ranking with retriever weights ($W_{Sparse}, \ W_{Dense} \ | \ W \in [0, 1]; \ W_{Sparse} + W_{Dense} = 1$), with higher scores indicating greater relevance.

\begin{equation}
    r_i = W_{Sparse} \times S(c_{i\_Sparse}) + W_{Dense} \times S(c_{i\_Dense})
\end{equation}

Then, we sort the obtained rank score with respect to $Q_R$ in descending order using the following equation, where, $\lambda$ is a constant to avoid division by 0. Finally, top $k$ chunks are retrieved.

\begin{equation}
    sort(Q_R, c_{i\_Retrieved}) = \sum_{i=1}^{\mathbb{N}} \frac{1}{\lambda + r_i(Q_R, c_{i\_Retrieved})}
\end{equation}

\begin{equation}
    \{ c_{i\_Sorted} : c_{i\_Sorted} \in C | 1 \leq i \leq k \} = sort(Q_R, c_{i\_Retrieved})
\end{equation}





The final decoded output will include only those chunks that contain the subjective, objective, assessment, or plan components from the given transcript. By passing the transcripts through the retriever-based filtering, we effectively reduce the tokens that do not correspond to the generation of SOAP notes, thereby abstracting unnecessary information before sending it for inference. This reduction in tokens not only helps \framework avoid the token overflow problem but also mitigates the inference model from hallucination.





\subsection{Inference}
The inference module receives the patient context, derived from the final retrieved concatenated chunks $c_{i}$, along with an \emph{instruction} , and a \emph{prompt} ($Q_{\mathcal{PLM}_{FT}}$) as shown in Fig. \ref{fig:soap_summaries}. The instruction guides the language model in performing its task. In our case, we utilize Alpaca prompt, as shown in Fig. \ref{fig:alpaca_prompt} as our instruction. On the other hand, prompt directs a fine-tuned PLM to produce clinical summaries in a zero-shot setting. The fine-tuned generator processes the prompt, patient context, and instructions to generate a comprehensive clinical SOAP summary. In the following subsections we describe our fine-tuning approach and then detail summary generation.

\subsubsection{Fine-tuning}
In this work, we leverage PEFT \cite{houlsby2019parameter} approach to fine-tune a PLM for clinical summary generation. PEFT enables efficient fine-tuning with minimal resources and costs. Specifically, we adopt Low Rank Adaptation (LoRA) \cite{hu2021lora} method and load pre-trained models onto a GPU as quantized 4-bit weights. Our motivation for this approach is two-folds: first, to explore the \textbf{feasibility} of training a PLM, such as LLAMA-3, on a single consumer GPU with 24GB of memory (e.g., Nvidia 4090), and second, to assess the \textbf{effectiveness} of fine-tuned PLMs with 4-bit precision in accurately generating clinical summaries. Additionally, PEFT helps prevent \emph{catastrophic forgetting}\cite{kirkpatrick2017overcoming} after the model has been trained \cite{meta}. We use the Alpaca prompt \cite{alpaca} for both fine-tuning and inference tasks, as illustrated in Fig. \ref{fig:alpaca_prompt}. The training is conducted using Supervised Fine-Tuning (SFT). More information on training dataset is provided in Section \ref{data_section}.

\begin{figure}[ht]
    \centering
    \begin{minipage}{8.8cm}
        \begin{tcolorbox}[enhanced,attach boxed title to top center={yshift=-1mm,yshifttext=-1mm},
            colback=green!12!white,colframe=gray!90!black,colbacktitle=gray!80!black, left=0.1mm, right=0.5mm, boxrule=0.50pt]
            \footnotesize
            {\fontfamily{cmr}\selectfont
            \textbf{Instruction}: """Below is an instruction that describes a task, paired with an input that provides further context. Write a response that appropriately completes the request. \\ \\
            \#\#\# Instruction: \\
                \{\}\\

            \#\#\# Input:\\
            \{\}\\

            \#\#\# Response:\\
            \{\}"""
            }
        \end{tcolorbox}
    \end{minipage}
    \vspace{-1mm}
    \caption{An example of an Alpaca prompt.}
    \label{fig:alpaca_prompt}
    \vspace{-3mm}
\end{figure}

\subsubsection{Summary Generation}

We utilize the output of the first module—the context, i.e., the decoded relevant chunks containing subjective, objective, assessment, and plan information from a given transcript—along with an instruction and a prompt as input to the inference module to generate a clinical summary. These input are concatenated and passed together to a fine-tuned PLM for summary generation, where $\mathcal{PLM}_{FT}$ is a fine-tuned PLM that understand how to generate a clinical summaries, $Q_{\mathcal{PLM}_{FT}}$ is a prompt (query), \emph{$inst$} denotes {instruction}  and [.,.,.] stands for concatenation.  

\begin{equation}
    \begin{aligned}
        Summary = \mathcal{PLM}_{FT}([context, prompt, inst]) = \\ 
        \mathcal{PLM}_{FT}([c_{i\_Sorted}, Q_{\mathcal{PLM}_{FT}}, inst])
    \end{aligned}
\end{equation}

Inference is conducted in a zero-shot setting using a fine-tuned PLM. The fine-tuning process equips the model to generalize effectively, allowing it to generate accurate clinical summaries even for new, unseen patient conversations without requiring additional few-shot examples. An example of the final clinical summary for a specific patient-doctor conversation is provided in Fig. \ref{fig:soap_summaries} (B) and (C) respectively.  





\section{Experiment \& Evaluation}
\label{experiment}

\subsection{Dataset Description and Preparation}
\label{data_section}

In this research, we utilize two different datasets for the fine-tuning task. The first dataset is the Figshare dataset, which contains 272 patient-doctor conversations. These conversations span five medical specialties: \emph{Cardiovascular, Gastrointestinal, Musculoskeletal, Dermatological}, and \emph{Respiratory}. Table \ref{example_figshare_data} provides an example from this dataset. The second dataset we use is the MTS-dialog dataset \cite{mts-dialog}, which contains 1,701 patient-doctor conversations. These conversations are centered around \emph{General Medicine, Orthopedic, Dermatology, Neurology}, and \emph{Allergy/Immunology}. From this dataset, we selected a subset of 1,201 clean conversations for our study. We then combined them, resulting in a total of 1,473 conversations. Additional statistics, including the total number of sentences, words, characters, unique vocabulary, and tokens for the conversations in the combined dataset are presented in Table \ref{figshare_and_MTS_data_stats}.
\begin{table}[ht]
\centering
\caption{Example of a doctor-patient conversation from the FigShare dataset.}
\label{example_figshare_data}
\renewcommand{\arraystretch}{1.10}%
\begin{tabular}{lll}
\cline{1-1}
\textbf{Patient-Doctor Conversation} &  &  \\ \cline{1-1}
\begin{tabular}[c]{@{}l@{}}D: What brought you in today?\\ P: Sure, I'm I'm just having a lot of chest pain and and so I thought I \\should get it checked out.\\ D: OK, before we start, could you remind me of your gender and age? \\ P: Sure 39, I'm a male.\\ D: OK, and so when did this chest pain start?\\ P: It started last night, but it's becoming sharper.\\ D: ...\\ P:...\end{tabular} &  &  \\ \cline{1-1}
\textbf{Medical   Speciality:} Cardiovascular                                                                                                                                                                                                                                                                                                                                                                &  &  \\ \cline{1-1}
\end{tabular}
\end{table}

\subsubsection{Ground-truth Generation} To fine-tune our models, we require ground truth data, which neither of these datasets provided. The MTS-dialog dataset contains very brief summaries, averaging less than three sentences, which are insufficient for our task. The Figshare dataset includes only conversations, with no summaries available. One key contribution of this paper is the creation of ground truth summaries for these conversations. To accomplish this, we collaborated closely with Potentia Analytics, a healthcare-focused data analytics and information technology company. We initially generated clinical summaries for all 1,473 conversations using the \emph{GPT-4-O-Mini} model (managed through API calls) in a zero-shot setting. Subject Matter Experts (SMEs), specifically medical doctors from Potentia Analytics, then manually evaluated and verified the factual correctness and contextual relevance of these summaries. 
Based on their feedback, we rectified any inconsistencies, discrepancies, or inaccuracies pertaining the summaries. We then created a final training dataset of 1,473 conversation-summary pairs, which we then utilized for our fine-tuning task. This dataset is publicly accessible on HuggingFace. Table \ref{ground_truth_data_stats} provides additional statistics for the ground-truth summaries, broken down into subjective, objective, assessment, and plan components, detailing the number of sentences, words, characters, vocabulary, and tokens.

\begin{table}[]
\centering
\caption{Statistics of the FigShare and MTS-Dialog datasets.}
\label{figshare_and_MTS_data_stats}
\renewcommand{\arraystretch}{1.30}%
\begin{tabular}{llllll}
\hline
\multicolumn{6}{c}{\textbf{Patient-Doctor Conversation (Figshare)}}                                                                                                                                           \\ \hline
\multicolumn{1}{l|}{\textbf{Metric}} & \multicolumn{1}{l|}{\textbf{Sentences}} & \multicolumn{1}{l|}{\textbf{Words}} & \multicolumn{1}{l|}{\textbf{Char}} & \multicolumn{1}{l|}{\textbf{Vocab}} & \textbf{Tokens} \\ \hline
\multicolumn{1}{l|}{Count}           & \multicolumn{1}{l|}{37910}              & \multicolumn{1}{l|}{369552}         & \multicolumn{1}{l|}{1478738}       & \multicolumn{1}{l|}{98535}          & 472384          \\ \hline
\multicolumn{1}{l|}{Mean}            & \multicolumn{1}{l|}{139.37}             & \multicolumn{1}{l|}{1358.64}        & \multicolumn{1}{l|}{5436.53}       & \multicolumn{1}{l|}{362.26}         & 1736.70         \\ \hline
\multicolumn{1}{l|}{Max}             & \multicolumn{1}{l|}{255}                & \multicolumn{1}{l|}{2401}           & \multicolumn{1}{l|}{9636}          & \multicolumn{1}{l|}{589}            & 3102            \\ \hline
\multicolumn{1}{l|}{Min}             & \multicolumn{1}{l|}{74}                 & \multicolumn{1}{l|}{808}            & \multicolumn{1}{l|}{3229}          & \multicolumn{1}{l|}{254}            & 1020            \\ \hline
\multicolumn{6}{c}{\textbf{Patient-Doctor   Conversation (MTS-Dialog)}}                                                                                                                                         \\ \hline
\multicolumn{1}{l|}{Count}           & \multicolumn{1}{l|}{15839}              & \multicolumn{1}{l|}{118558}         & \multicolumn{1}{l|}{500393}        & \multicolumn{1}{l|}{72225}          & 152232          \\ \hline
\multicolumn{1}{l|}{Mean}            & \multicolumn{1}{l|}{13.18}              & \multicolumn{1}{l|}{98.71}          & \multicolumn{1}{l|}{416.64}        & \multicolumn{1}{l|}{60.13}          & 126.75          \\ \hline
\multicolumn{1}{l|}{Max}             & \multicolumn{1}{l|}{167}                & \multicolumn{1}{l|}{1474}           & \multicolumn{1}{l|}{6823}          & \multicolumn{1}{l|}{457}            & 2038            \\ \hline
\multicolumn{1}{l|}{Min}             & \multicolumn{1}{l|}{1}                  & \multicolumn{1}{l|}{1}              & \multicolumn{1}{l|}{8}             & \multicolumn{1}{l|}{1}              & 3               \\ \hline
\end{tabular}
\end{table}

\begin{table*}[ht]
\centering
\caption{Statistics of ground-truth clinical summaries from 1,473 patient-doctor conversations.}
\label{ground_truth_data_stats}
\renewcommand{\arraystretch}{1.30}%
\begin{tabular}{lllllllllll}
\hline
\multicolumn{11}{c}{\textbf{Clinical Summaries Statistics}}                                                                                                                                                                                                                                                                                              \\ \hline
\multicolumn{1}{l|}{\multirow{2}{*}{\textbf{Metric}}} & \multicolumn{2}{c|}{\textbf{Subjective}}                              & \multicolumn{2}{c|}{\textbf{Objective}}                               & \multicolumn{2}{c|}{\textbf{Assessment}}                               & \multicolumn{2}{c|}{\textbf{Plan}}                                    & \multicolumn{2}{c}{\textbf{Vocab/Token}}            \\ \cline{2-11} 
\multicolumn{1}{l|}{}                        & \multicolumn{1}{l|}{Sentences} & \multicolumn{1}{l|}{Words}  & \multicolumn{1}{l|}{Sentences} & \multicolumn{1}{l|}{Words}  & \multicolumn{1}{l|}{Sentences} & \multicolumn{1}{l|}{Words}  & \multicolumn{1}{l|}{Sentences} & \multicolumn{1}{l|}{Words}  & \multicolumn{1}{l|}{Vocab}  & Tokens  \\ \hline
\multicolumn{1}{l|}{Count}                   & \multicolumn{1}{l|}{48775}     & \multicolumn{1}{l|}{44927} & \multicolumn{1}{l|}{1957}      & \multicolumn{1}{l|}{12330} & \multicolumn{1}{l|}{1912}      & \multicolumn{1}{l|}{11494} & \multicolumn{1}{l|}{2017}      & \multicolumn{1}{l|}{19396} & \multicolumn{1}{l|}{77384} & 226246 \\ \hline
\multicolumn{1}{l|}{Max}                     & \multicolumn{1}{l|}{22}        & \multicolumn{1}{l|}{323}    & \multicolumn{1}{l|}{9}         & \multicolumn{1}{l|}{110}    & \multicolumn{1}{l|}{7}         & \multicolumn{1}{l|}{110}    & \multicolumn{1}{l|}{6}         & \multicolumn{1}{l|}{80}     & \multicolumn{1}{l|}{198}    & 500     \\ \hline
\multicolumn{1}{l|}{Mean}                    & \multicolumn{1}{l|}{3.25}      & \multicolumn{1}{l|}{63.43}  & \multicolumn{1}{l|}{1.328}     & \multicolumn{1}{l|}{8.37}   & \multicolumn{1}{l|}{1.298}     & \multicolumn{1}{l|}{10.02}  & \multicolumn{1}{l|}{1.371}     & \multicolumn{1}{l|}{13.17}  & \multicolumn{1}{l|}{18.85}  & 36.47   \\ \hline
\end{tabular}
\end{table*}


\begin{table}[ht]
\centering
\caption{Evaluation dataset statistics from staged patient-doctor conversations.}
\label{staged_conversation_stats}
\renewcommand{\arraystretch}{1.30}%
\begin{tabular}{llllll}
\hline
\multicolumn{6}{c}{\textbf{Staged Conversation Statistics}}                                                                                                                                                    \\ \hline
\multicolumn{1}{l|}{\textbf{Metric}} & \multicolumn{1}{l|}{\textbf{Sentences}} & \multicolumn{1}{l|}{\textbf{Words}} & \multicolumn{1}{l|}{\textbf{Char}} & \multicolumn{1}{l|}{\textbf{Vocab}} & \textbf{Tokens} \\ \hline
\multicolumn{1}{l|}{Count}           & \multicolumn{1}{l|}{1997}               & \multicolumn{1}{l|}{20359}          & \multicolumn{1}{l|}{99518}         & \multicolumn{1}{l|}{6965}           & 25690           \\ \hline
\multicolumn{1}{l|}{Mean}            & \multicolumn{1}{l|}{99.85}              & \multicolumn{1}{l|}{1017.95}        & \multicolumn{1}{l|}{4975.9}        & \multicolumn{1}{l|}{348.25}         & 1284.5          \\ \hline
\multicolumn{1}{l|}{Max}             & \multicolumn{1}{l|}{146}                & \multicolumn{1}{l|}{1573}           & \multicolumn{1}{l|}{7886}          & \multicolumn{1}{l|}{448}            & 1995            \\ \hline
\multicolumn{1}{l|}{Min}             & \multicolumn{1}{l|}{61}                 & \multicolumn{1}{l|}{551}            & \multicolumn{1}{l|}{2629}          & \multicolumn{1}{l|}{222}            & 708             \\ \hline
\end{tabular}
\end{table}

\begin{table*}[ht]
\centering
\caption{Comparison of GPT models using zero-shot prompting and \framework for generating clinical summaries, evaluated with lexical-based (ROUGE) and embedding-based (BERTScore) metrics.}
\label{quant_bert_and_rogue}
\renewcommand{\arraystretch}{1.30}%
\label{tab:my-table}
\begin{tabular}{l|lll|lll|lll|lll}
\hline
   &
  \multicolumn{3}{l|}{\textbf{Rouge-1}} &
  \multicolumn{3}{l|}{\textbf{Rouge-2}} &
  \multicolumn{3}{l|}{\textbf{Rouge-L}} &
  \multicolumn{3}{l}{\textbf{BertScore}} \\ \cline{2-13} 
  \textbf{Model} &
  \multicolumn{1}{l|}{\textbf{P}} &
  \multicolumn{1}{l|}{\textbf{R}} &
  \textbf{F-1} &
  \multicolumn{1}{l|}{\textbf{P}} &
  \multicolumn{1}{l|}{\textbf{R}} &
  \textbf{F-1} &
  \multicolumn{1}{l|}{\textbf{P}} &
  \multicolumn{1}{l|}{\textbf{R}} &
  \textbf{F-1} &
  \multicolumn{1}{l|}{\textbf{P}} &
  \multicolumn{1}{l|}{\textbf{R}} &
  \textbf{F-1} \\ \hline
  GPT-4-Turbo &
  \multicolumn{1}{l|}{\cellcolor[HTML]{FD6864}0.50} &
  \multicolumn{1}{l|}{\cellcolor[HTML]{34FF34}0.72} &
  \cellcolor[HTML]{FD6864}0.58 &
  \multicolumn{1}{l|}{\cellcolor[HTML]{FD6864}0.24} &
  \multicolumn{1}{l|}{\cellcolor[HTML]{FFCCC9}0.35} &
  \cellcolor[HTML]{FD6864}0.29 &
  \multicolumn{1}{l|}{\cellcolor[HTML]{FD6864}0.31} &
  \multicolumn{1}{l|}{\cellcolor[HTML]{FFCCC9}0.45} &
  \cellcolor[HTML]{FD6864}0.36 &
  \multicolumn{1}{l|}{\cellcolor[HTML]{FD6864}0.73} &
  \multicolumn{1}{l|}{\cellcolor[HTML]{FD6864}0.74} &
  \cellcolor[HTML]{FD6864}0.73 \\ \cline{1-13} 
  GPT-4-0-Mini &
  \multicolumn{1}{l|}{0.64} &
  \multicolumn{1}{l|}{0.66} &
  \cellcolor[HTML]{FFCCC9}0.64 &
  \multicolumn{1}{l|}{0.38} &
  \multicolumn{1}{l|}{0.39} &
  0.38 &
  \multicolumn{1}{l|}{0.45} &
  \multicolumn{1}{l|}{0.46} &
  0.45 &
  \multicolumn{1}{l|}{0.76} &
  \multicolumn{1}{l|}{\cellcolor[HTML]{FFCCC9}0.78} &
  \cellcolor[HTML]{FFCCC9}0.77 \\ \cline{1-13} 
  GPT-3.5-Turbo &
  \multicolumn{1}{l|}{0.64} &
  \multicolumn{1}{l|}{0.59} &
  \cellcolor[HTML]{FFCCC9}0.61 &
  \multicolumn{1}{l|}{0.35} &
  \multicolumn{1}{l|}{\cellcolor[HTML]{FD6864}0.32} &
  0.33 &
  \multicolumn{1}{l|}{0.42} &
  \multicolumn{1}{l|}{\cellcolor[HTML]{FD6864}0.39} &
  \cellcolor[HTML]{FFCCC9}0.40 &
  \multicolumn{1}{l|}{\cellcolor[HTML]{FD6864}0.74} &
  \multicolumn{1}{l|}{0.79} &
  \cellcolor[HTML]{FFCCC9}0.76 \\ \hline
  \framework\textsubscript{LLAMA-3-8B} &
  \multicolumn{1}{l|}{\cellcolor[HTML]{34FF34}0.72} &
  \multicolumn{1}{l|}{\cellcolor[HTML]{9AFF99}0.69} &
  \cellcolor[HTML]{34FF34}0.70 &
  \multicolumn{1}{l|}{\cellcolor[HTML]{34FF34}0.50} &
  \multicolumn{1}{l|}{\cellcolor[HTML]{34FF34}0.48} &
  \cellcolor[HTML]{34FF34}0.48 &
  \multicolumn{1}{l|}{\cellcolor[HTML]{34FF34}0.57} &
  \multicolumn{1}{l|}{\cellcolor[HTML]{34FF34}0.54} &
  \cellcolor[HTML]{34FF34}0.55 &
  \multicolumn{1}{l|}{\cellcolor[HTML]{34FF34}0.87} &
  \multicolumn{1}{l|}{\cellcolor[HTML]{67FD9A}0.82} &
  \cellcolor[HTML]{34FF34}0.84 \\ \cline{1-13} 

  \framework\textsubscript{Mistral-Nemo-12B} &
  \multicolumn{1}{l|}{0.67} &
  \multicolumn{1}{l|}{\cellcolor[HTML]{67FD9A}0.72} &
  \cellcolor[HTML]{67FD9A}0.68 &
  \multicolumn{1}{l|}{\cellcolor[HTML]{9AFF99}0.44} &
  \multicolumn{1}{l|}{\cellcolor[HTML]{34FF34}0.48} &
  \cellcolor[HTML]{9AFF99}0.45 &
  \multicolumn{1}{l|}{\cellcolor[HTML]{FFFFFF}0.50} &
  \multicolumn{1}{l|}{\cellcolor[HTML]{34FF34}0.54} &
  \cellcolor[HTML]{9AFF99}0.51 &
  \multicolumn{1}{l|}{0.76} &
  \multicolumn{1}{l|}{\cellcolor[HTML]{67FD9A}0.82} &
  0.78 \\ \cline{1-13} 

  \framework\textsubscript{Mistral-7B} &
  \multicolumn{1}{l|}{\cellcolor[HTML]{34FF34}0.70} &
  \multicolumn{1}{l|}{\cellcolor[HTML]{9AFF99}0.69} &
  \cellcolor[HTML]{67FD9A}0.68 &
  \multicolumn{1}{l|}{\cellcolor[HTML]{67FD9A}0.46} &
  \multicolumn{1}{l|}{0.46} &
  \cellcolor[HTML]{9AFF99}0.45 &
  \multicolumn{1}{l|}{\cellcolor[HTML]{9AFF99}0.51} &
  \multicolumn{1}{l|}{\cellcolor[HTML]{9AFF99}0.50} &
  0.49 &
  \multicolumn{1}{l|}{\cellcolor[HTML]{FFCCC9}0.75} &
  \multicolumn{1}{l|}{\cellcolor[HTML]{34FF34}0.83} &
  0.79 \\ \cline{1-13} 
  \framework\textsubscript{Gemma-2-9B} &
  \multicolumn{1}{l|}{\cellcolor[HTML]{9AFF99}0.69} &
  \multicolumn{1}{l|}{0.67} &
  0.67 &
  \multicolumn{1}{l|}{\cellcolor[HTML]{9AFF99}0.44} &
  \multicolumn{1}{l|}{0.43} &
  0.43 &
  \multicolumn{1}{l|}{\cellcolor[HTML]{FFFFFF}0.50} &
  \multicolumn{1}{l|}{0.48} &
  0.49 &
  \multicolumn{1}{l|}{\cellcolor[HTML]{67FD9A}0.82} &
  \multicolumn{1}{l|}{\cellcolor[HTML]{34FF34}0.83} &
  \cellcolor[HTML]{67FD9A}0.82 \\ \hline
\end{tabular}%
\end{table*}

\subsection{Evaluation}
\label{evaluation}

Due to the strict privacy concerns and Health Insurance Portability and Accountability Act (HIPAA) regulations around patient data, coupled with its inaccessibility, we opted to simulate patient-doctor conversations. For this, we partnered with the Department of Theatre \& Film at Mississippi State University to create 20 simulated conversations. These conversations were staged as role-playing scenarios, where theater arts students simulated realistic interactions between patients and doctors. The conversations were recorded in WAV format and subsequently processed using Automatic Speech Recognition (ASR) techniques, specifically utilizing the Whisper-large\footnote{https://huggingface.co/openai/whisper-large} model. The average length of these role-played conversations is approximately 9 minutes. Additional statistics on these conversations are provided in Table \ref{staged_conversation_stats}. We then utilized these simulated conversations as our evaluation dataset to assess the robustness of \framework for the clinical summaries generation task. The evaluation was conducted through both automatic and manual methods.

In the following subsections, we discuss the results of our assessment.

\subsubsection{Automatic Evaluation}
\label{quantitative_evalaution}


In this paper, we consider two types of metrics: \emph{lexical-based} and \emph{text-embedding-based}, to evaluate the clinical summaries generated by \framework. For the lexical-based metric, we choose Recall-Oriented Understudy for Gisting Evaluation (ROUGE)\cite{lin2004rouge}, which primarily focuses on lexical overlaps between generated summaries and the ground truth but does not capture the semantic meaning of the summaries. Considering the limitation of ROUGE, we also employ text-embedding-based metrics, such as BertScore \cite{zhang2019bertscore}. It uses pre-trained contextual embeddings from a BERT-based model to evaluate the semantic similarity between the ground truth and generated summaries by computing cosine similarity. Specifically, we utilize the \emph{deberta-xlarge-mnli}\footnote{https://huggingface.co/microsoft/deberta-xlarge-mnli} model in our experiments.


Table \ref{quant_bert_and_rogue} presents the clinical summarization results. We compared the performance of the state-of-the-art GPT-based proprietary PLMs with our approach (combination of retriever-based filtering and fine-tuning) equipped with open source models in generating clinical summaries from patient-doctor conversation in zero-shot settings. Notably, \framework, particularly when paired with LLAMA-3, outperformed GPT-based models in both ROUGE and BERTScore metrics.

In terms of ROUGE-1, which measures unigram overlap, \emph{LLAMA 3-8B} achieved the highest precision of 0.72 and F1-score of 0.70. A high precision score indicates that a large proportion of the words generated by models (unigrams) are also found in the ground-truth summary, whereas a high F-1 score reflects a model's overall effectiveness in producing a summary that is accurate and covers the ground truth well. In contrast, the \emph{GPT-4-Turbo} model scored the lowest with an F-1 score of 0.58 and precision of 0.50. This trend persisted in ROUGE-2, where \emph{LLAMA 3-8B} led with an F1-score of 0.48 and precision of 0.50, significantly outperforming \emph{GPT-4-Turbo}, which only achieved an F-1 score of 0.29 and precision of 0.24. Similarly, for ROUGE-L, which assesses the longest common sub-sequence between generated and reference texts, \emph{LLAMA 3-8B} excelled with an F1-score of 0.55 and precision of 0.48, while \emph{GPT-4-Turbo} lagged behind with an F1-score of 0.36 and precision of 0.31. BERTScore, which evaluates the semantic similarity between generated and ground-turth summaries, further corroborated these findings. \emph{LLAMA 3-8B} stood out with the highest F-1 of 0.84 and precision on 0.87, reflecting its strong alignment in meaning with the ground truth summary. Conversely, \emph{GPT-4-Turbo} recorded the lowest F-1 of 0.73, indicating its relative difficulty in generating semantically accurate summaries. The second best performing model with our approach is \emph{Gemma-2-9B} with impressive F-1 score of 0.82, and precision of 0.82.

\subsubsection{Expert Human Evaluation}
\label{qualitative_evaluation}

Expert human evaluation plays a critical role in assessing the quality of generated summaries, especially as automatic metrics like ROUGE and BERTScore, though useful, may not always align with expert judgment \cite{deutsch2022re,giorgi2023wanglab}. Recognizing these limitations, we incorporated human evaluation in this study. Given the expensive nature of human evaluation, we assembled a panel of four SMEs, including medical resident doctors and physicians, to compare the summaries generated by \framework with those from GPT-based models with prompting techniques. Specifically, we focused on the best-performing models from both systems for zero-shot summarization. Using the same set of 20 conversations and summaries as in the quantitative analysis ensured a fair comparison.

\begin{figure}
  \centering
  \includegraphics[width=.95\linewidth]{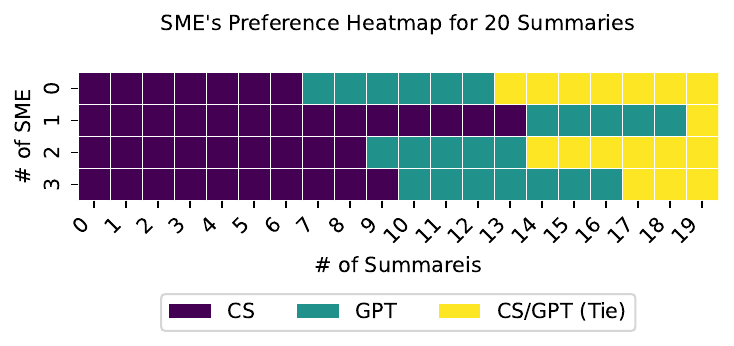}
  \caption{Heatmap illustrating the preferences between summaries generated by \framework and GPT, along with ties indicating equal preference between the two.}
  \label{fig:heat_map_agreement}
\end{figure}

\begin{table}[ht]
\caption{An overview of human evalaution. Four SME's evaluated summaries generated by \framework (CS) and GPT-O-Mini (GPT). For each case, the SMEs selected their preferred summary. The win rate represents the percentage of cases where a summary was preferred, with ties excluded from the calculation.}
\label{tab:human_eval}
\renewcommand{\arraystretch}{1.20}%
\centering
\begin{tabular}{c|ll|l|ll}
\hline
\multicolumn{1}{l|}{\multirow{2}{*}{\textbf{SME}}} & \multicolumn{2}{l|}{\textbf{Preferred}}         & \textbf{Ties}    & \multicolumn{2}{l}{\textbf{Win rate \%}}          \\ \cline{2-6} 
\multicolumn{1}{l|}{}                     & \multicolumn{1}{l|}{\textbf{CS}} & \textbf{GPT} & \textbf{CS/ GPT} & \multicolumn{1}{l|}{\textbf{CS}} & \textbf{GPT} \\ \hline
1                                          & \multicolumn{1}{l|}{7}           & 6            & 7                & \multicolumn{1}{l|}{0.54}        & 0.46         \\ \hline
2                                          & \multicolumn{1}{l|}{14}          & 5            & 1                & \multicolumn{1}{l|}{0.74}        & 0.26         \\ \hline
3                                          & \multicolumn{1}{l|}{9}           & 5            & 6                & \multicolumn{1}{l|}{0.64}        & 0.36         \\ \hline
4                                          & \multicolumn{1}{l|}{9}           & 8            & 3                & \multicolumn{1}{l|}{0.53}        & 0.47         \\ \hline
\multicolumn{1}{l|}{\textbf{Total}}       & \multicolumn{1}{l|}{39}          & 24           & 17               & \multicolumn{1}{l|}{\cellcolor{green!15}0.61}        & 0.39         \\ \hline
\end{tabular}
\end{table}
Following the evaluation strategy outlined by Giorgi et al. \cite{giorgi2023wanglab}, the SMEs were provided with ground-truth data, summaries from LLAMA-3-8B (best performing model) in our framework, and summaries from GPT-4-O-Mini (best performing among gpt models). Summaries were anonymized and labeled as `A' and `B' with the SMEs instructed to choose their preferred version or select both if there is a tie. While our evaluation strategy is similar to \cite{giorgi2023wanglab}, there are some key differences. They relied on three criteria: \emph{critical, non-critical}, and \emph{irrelevant information} from previous research by Savkov et al.\cite{savkov2022consultation} to guide SME preferences. Additionally, they included the ground truth in their evaluation, while we focused solely on comparing the summaries generated by \framework and those from GPT-based models. In contrast, we introduced a fourth criterion: \emph{factual correctness} (must capture all key factual information). Based on this, we redefine a good summary as one that is \emph{``factually accurate, includes all critical information, some non-critical information, and contains minimal irrelevant details''}. 
The results of this evaluation are shown in Table \ref{tab:human_eval}, and we visualize the agreement between SMEs using heatmaps, as illustrated in Fig. \ref{fig:heat_map_agreement}.

Overall, the summaries generated by \framework are strongly preferred over summaries generated by the GPT-based models further validating the high performance reported by the automatic evaluation metrics. In addition, we assessed Inter-Rater Reliability (IRR) among the four SMEs using two statistical measures: Fleiss' Kappa ($\kappa$) \cite{mchugh2012interrater} and Krippendorff's Alpha ($\alpha$). The results, shown in Table \ref{tab:felis}, indicate moderate agreement (0.41 to 0.60) for both $\kappa$ and $\alpha$. This suggests that while the SMEs were not perfectly aligned in their preferences, they demonstrated a fair level of consensus. In our opinion, this variability arises from differences in the SMEs' experience levels and subjective interpretations, which align with similarly low agreement scores reported in previous research \cite{neupane2024medinsight, giorgi2023wanglab, zhang2021leveraging}. 

\begin{table}
\caption{IRR metrics, including Fleiss' Kappa ($\kappa$) and Krippendorff's Alpha ($\alpha$), demonstrate moderate agreement among the four SMEs.} \label{tab:felis}
    \centering
    
    {\renewcommand{\arraystretch}{1.3}%
\footnotesize

\begin{tabular}{ll|l} 
\hline
                                    & \multicolumn{2}{l}{\textbf{Inter-Rater Reliability (IRR)}}  \\ \cline{1-3}
\multirow{2}{*}{Overall Agreement}  & Fleiss Kappa  ($\kappa$)             & Krippendorff's Alpha $(\alpha)$      \\ \cline{2-3}
                                    &  0.43746         & \cellcolor{green!15} 0.44450          \\ \cline{1-3}
\end{tabular}

}

\end{table}

\begin{figure}[ht]
  \centering
  \includegraphics[width=0.8\linewidth]{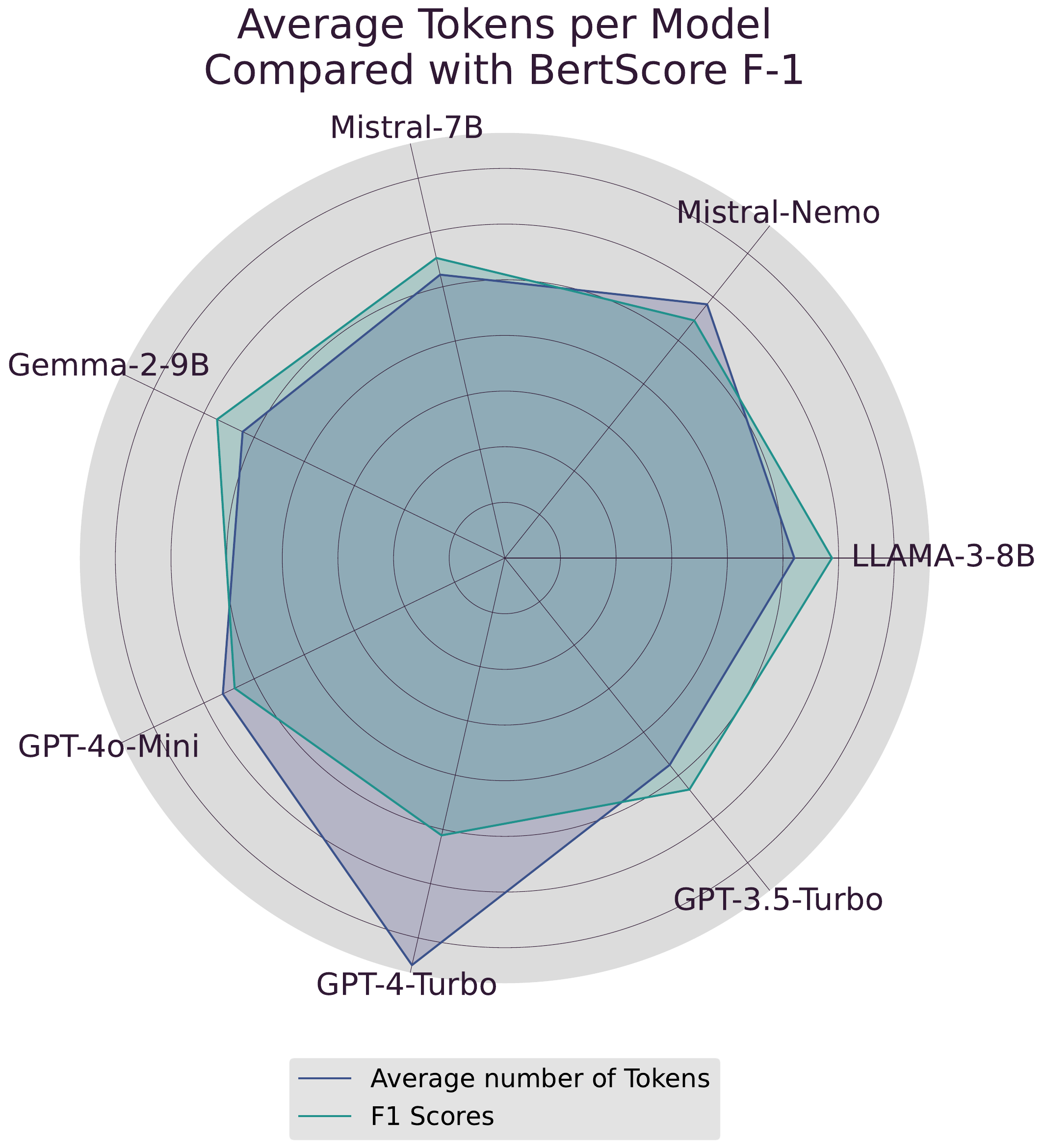}
  \caption{Radar chart illustrating how different models compare in terms of two key metrics: the average number of tokens and F-1 scores of BertScore.}
  \label{fig:radar_avg_vs_f1}
  \vspace{-3mm}
\end{figure}

\noindent \textbf{Findings with respect to \# number of Tokens:}
In this study, we further investigated whether there is a quantitative correlation between the best and worst performing models in terms of F-1 score and their average token count. Fig. \ref{fig:radar_avg_vs_f1} presents a radar chart comparing the average tokens per model with their corresponding BERTScore F-1 scores. 
Token counts for both GPT-based models and open-source PLMs were computed using the \emph{BAAI/bge-large-en-v1.5} model from Hugging Face. 
A key finding from this analysis is the correlation between token count variability and accuracy: LLAMA-3-8B (avg token count: 260) and Gemma-2-9B (average token count: 262), which closely matches the ground truth (avg token 268), achieves the highest BERTScore, while GPT-4-Turbo (avg token count: 375), with significantly high token usage likely hallucinating, performs the worst. Mistral-Nemo-12B (average token count:292) and GPT-4-0-Mini (average token count: 281) also show increased token generation in some cases but have a tendency to generate fewer tokens, while Mistral-7B (average token count: 259) and GPT-3.5-Turbo (average token count: 237) generally produce fewer tokens, potentially missing critical details. Despite these variations, LLAMA-3-8B and Gemma-2-9B maintain a balanced approach of using and generating tokens that are strictly coherent with the context provided to them by retriever-based filtering, producing summaries that are aligned with the ground truth.

This suggests that models generating summaries with token counts closer to the ground truth tend to produce more accurate outputs. In contrast, greater variability in token count, as seen in GPT-4-Turbo and GPT-3.5-Turbo, may lead to less accurate summaries. This conclusion is further supported by expert human assessment, where SMEs showed a clear preference for the summaries generated by LLAMA-3-8B over those produced by GPT-4-Turbo.


\label{results}

\section{Limitations and Discussion}
\label{limitation}
Despite \framework's encouraging results in generating good clinical summaries, several limitations should be noted. First, the model’s performance is highly reliant on the quality and diversity of the training data used in fine-tuning task. The fine-tuning dataset, comprising 1,473 conversations from the FigShare and MTS-Dialog datasets with clinical summaries generated by a PLM and validated by SMEs, is limited in scope, as it only encompasses a narrow range of medical specialties.

Another limitation is the use of simulated patient-doctor conversations in the evaluation phase. While these simulated conversations are useful for HIPAA compliance, they may fail to capture the complexities and variability of real-world clinical interactions. Consequently, the generated summaries may not perform as well in real-world clinical settings, where patient communication is less structured and more nuanced.

Furthermore, while the framework uses retrieval-based filtering to improve factual accuracy, the risk of hallucinations persists, especially when summarizing conversations with ambiguous or incomplete information. Ensuring factual accuracy is critical in medical settings, and additional validation mechanisms may be necessary to mitigate these risks. Moreover, while we added a criterion of factual correctness to the human evaluation process, the moderate IRR scores indicate that subjective interpretation among SMEs can still result in inconsistencies in summary evaluations. Another important aspect is Biases, PLMs trained on vast amounts of text data may inadvertently capture and reproduce biases present in the data. For example, it may over-prioritize common condition such as ``upper respiratory infections'' when interpreting symptoms, potentially overlooking rarer but more serious conditions.

Lastly, the \framework's dependence on fine-tuning PLM necessitates substantial computational resources, potentially constraining its scalability in low-resource clinical settings. The current implementation is tailored to operate on a consumer GPU with 24GB of memory, which may not be readily available to all healthcare institutions. Due to limited computational resources, we focused our efforts on models with $\leq$ 12 Billion parameters. Fine-tuning a bigger PLM for example model with 30B, 70B or higher parameter may yield even better results.

\section{Conclusion \& Future Work}
\label{conclusion}

In this paper, we demonstrated the feasibility of automatically generating clinical summaries directly from patient-doctor conversations using a framework with two module architecture referred to as \framework. The first module, retriever-based filtering, acts as an extractive component, identifying relevant portions of the transcript that contain subjective, objective, assessment, and plan information. The advantage of this approach is that it not only filters out unnecessary information from the transcripts but also reduces the risk of hallucination by passing only the relevant chunks to the second module. The second module, inference, utilizes the filtered information as context and uses a fine-tuned PLM to generate clinical summaries through abstraction.

We created a high-quality fine-tuning training dataset consisting of 1,473 conversation-summary pairs and used it to fine-tune four open-source PLMs with $\leq$ 12B parameters. Surprisingly, when combined with our framework for inference, these fine-tuned open-source PLMs substantially outperformed state-of-the-art GPT models in both automatic and expert human evaluations. Expert human assessments by SMEs confirmed that the summaries generated by \framework were more preferable than those produced by GPT models using prompting. We believe our results are encouraging, and \framework offers a promising solution for automating clinical summarization.

Future work will focus on expanding the both training and validation dataset, improving framework's scalability, and exploring real-world applications in diverse clinical settings. We also intend to further investigate methods to further reduce hallucination and potential biases of PLMs.







\section*{Acknowledgement}
\label{ack}
The research reported in this publication is jointly supported by the Predictive Analytics and Technology Integration (PATENT) Laboratory in the Department of Computer Science and Engineering at Mississippi State University, Potentia Analytics, Inc., and partially by the National Institute of Nursing Research of the National Institutes of Health under Award Number R41NR021089. The content is solely the responsibility of the authors and does not necessarily represent the official views of the National Institutes of Health. We also extend our gratitude to Dr. Nina Marhamati and Ivan A. Fernandez for their constructive inputs during the final stages of the paper.

\bibliographystyle{IEEEtran}
\bibliography{references}

\begin{thebibliography}{10}
\providecommand{\url}[1]{#1}
\csname url@samestyle\endcsname
\providecommand{\newblock}{\relax}
\providecommand{\bibinfo}[2]{#2}
\providecommand{\BIBentrySTDinterwordspacing}{\spaceskip=0pt\relax}
\providecommand{\BIBentryALTinterwordstretchfactor}{4}
\providecommand{\BIBentryALTinterwordspacing}{\spaceskip=\fontdimen2\font plus
\BIBentryALTinterwordstretchfactor\fontdimen3\font minus \fontdimen4\font\relax}
\providecommand{\BIBforeignlanguage}[2]{{%
\expandafter\ifx\csname l@#1\endcsname\relax
\typeout{** WARNING: IEEEtran.bst: No hyphenation pattern has been}%
\typeout{** loaded for the language `#1'. Using the pattern for}%
\typeout{** the default language instead.}%
\else
\language=\csname l@#1\endcsname
\fi
#2}}
\providecommand{\BIBdecl}{\relax}
\BIBdecl

\bibitem{neupane2024medinsight}
S.~Neupane, S.~Mitra, S.~Mittal, N.~A. Golilarz, S.~Rahimi, and A.~Amirlatifi, ``Medinsight: A multi-source context augmentation framework for generating patient-centric medical responses using large language models,'' \emph{arXiv preprint arXiv:2403.08607}, 2024.

\bibitem{zhao2024retrieval}
P.~Zhao, H.~Zhang, Q.~Yu, Z.~Wang, Y.~Geng, F.~Fu, L.~Yang, W.~Zhang, and B.~Cui, ``Retrieval-augmented generation for ai-generated content: A survey,'' \emph{arXiv preprint arXiv:2402.19473}, 2024.

\bibitem{mcguire1996remembering}
L.~C. Mcguire, ``Remembering what the doctor said: organization and adults' memory for medical information,'' \emph{Experimental aging research}, vol.~22, no.~4, pp. 403--428, 1996.

\bibitem{anderson1979patient}
J.~L. Anderson, S.~Dodman, M.~Kopelman, and A.~Fleming, ``Patient information recall in a rheumatology clinic,'' \emph{Rheumatology}, vol.~18, no.~1, pp. 18--18, 1979.

\bibitem{kumar2016burnout}
S.~Kumar, ``Burnout and doctors: prevalence, prevention and intervention,'' in \emph{Healthcare}, vol.~4, no.~3.\hskip 1em plus 0.5em minus 0.4em\relax MDPI, 2016, p.~37.

\bibitem{knoll2022user}
T.~Knoll, F.~Moramarco, A.~P. Korfiatis, R.~Young, C.~Ruffini, M.~Perera, C.~Perstl, E.~Reiter, A.~Belz, and A.~Savkov, ``User-driven research of medical note generation software,'' \emph{arXiv preprint arXiv:2205.02549}, 2022.

\bibitem{huang2023survey}
L.~Huang, W.~Yu, W.~Ma, W.~Zhong, Z.~Feng, H.~Wang, Q.~Chen, W.~Peng, X.~Feng, B.~Qin \emph{et~al.}, ``A survey on hallucination in large language models: Principles, taxonomy, challenges, and open questions,'' \emph{arXiv preprint arXiv:2311.05232}, 2023.

\bibitem{mts-dialog}
\BIBentryALTinterwordspacing
A.~Ben~Abacha, W.-w. Yim, Y.~Fan, and T.~Lin, ``An empirical study of clinical note generation from doctor-patient encounters,'' in \emph{Proceedings of the 17th Conference of the European Chapter of the Association for Computational Linguistics}.\hskip 1em plus 0.5em minus 0.4em\relax Dubrovnik, Croatia: Association for Computational Linguistics, May 2023, pp. 2291--2302. [Online]. Available: \url{https://aclanthology.org/2023.eacl-main.168}
\BIBentrySTDinterwordspacing

\bibitem{zhang2021leveraging}
L.~Zhang, R.~Negrinho, A.~Ghosh, V.~Jagannathan, H.~R. Hassanzadeh, T.~Schaaf, and M.~R. Gormley, ``Leveraging pretrained models for automatic summarization of doctor-patient conversations,'' \emph{arXiv preprint arXiv:2109.12174}, 2021.

\bibitem{giorgi2023wanglab}
J.~Giorgi, A.~Toma, R.~Xie, S.~S. Chen, K.~R. An, G.~X. Zheng, and B.~Wang, ``Wanglab at mediqa-chat 2023: Clinical note generation from doctor-patient conversations using large language models,'' \emph{arXiv preprint arXiv:2305.02220}, 2023.

\bibitem{brown2020language}
T.~Brown, B.~Mann, N.~Ryder, M.~Subbiah, J.~D. Kaplan, P.~Dhariwal, A.~Neelakantan, P.~Shyam, G.~Sastry, A.~Askell \emph{et~al.}, ``Language models are few-shot learners,'' \emph{Advances in neural information processing systems}, vol.~33, pp. 1877--1901, 2020.

\bibitem{schloss2020towards}
B.~Schloss and S.~Konam, ``Towards an automated soap note: classifying utterances from medical conversations,'' in \emph{Machine Learning for Healthcare Conference}.\hskip 1em plus 0.5em minus 0.4em\relax PMLR, 2020, pp. 610--631.

\bibitem{sinsky2016allocation}
C.~Sinsky, L.~Colligan, L.~Li, M.~Prgomet, S.~Reynolds, L.~Goeders, J.~Westbrook, M.~Tutty, and G.~Blike, ``Allocation of physician time in ambulatory practice: a time and motion study in 4 specialties,'' \emph{Annals of internal medicine}, vol. 165, no.~11, pp. 753--760, 2016.

\bibitem{weed1971problem}
L.~Weed, ``The problem oriented record as a basic tool in medical education, patient care and clinical research.'' \emph{Annals of clinical research}, vol.~3, no.~3, pp. 131--134, 1971.

\bibitem{vaswani2017attention}
A.~Vaswani, N.~Shazeer, N.~Parmar, J.~Uszkoreit, L.~Jones, A.~N. Gomez, {\L}.~Kaiser, and I.~Polosukhin, ``Attention is all you need,'' \emph{Advances in neural information processing systems}, vol.~30, 2017.

\bibitem{min2023recent}
B.~Min, H.~Ross, E.~Sulem, A.~P.~B. Veyseh, T.~H. Nguyen, O.~Sainz, E.~Agirre, I.~Heintz, and D.~Roth, ``Recent advances in natural language processing via large pre-trained language models: A survey,'' \emph{ACM Computing Surveys}, vol.~56, no.~2, pp. 1--40, 2023.

\bibitem{wang2023survey}
C.~Wang, X.~Liu, Y.~Yue, X.~Tang, T.~Zhang, C.~Jiayang, Y.~Yao, W.~Gao, X.~Hu, Z.~Qi \emph{et~al.}, ``Survey on factuality in large language models: Knowledge, retrieval and domain-specificity,'' \emph{arXiv preprint arXiv:2310.07521}, 2023.

\bibitem{kirkpatrick2017overcoming}
J.~Kirkpatrick, R.~Pascanu, N.~Rabinowitz, J.~Veness, G.~Desjardins, A.~A. Rusu, K.~Milan, J.~Quan, T.~Ramalho, A.~Grabska-Barwinska \emph{et~al.}, ``Overcoming catastrophic forgetting in neural networks,'' \emph{Proceedings of the national academy of sciences}, vol. 114, no.~13, pp. 3521--3526, 2017.

\bibitem{lewis2020retrieval}
P.~Lewis, E.~Perez, A.~Piktus, F.~Petroni, V.~Karpukhin, N.~Goyal, H.~K{\"u}ttler, M.~Lewis, W.-t. Yih, T.~Rockt{\"a}schel \emph{et~al.}, ``Retrieval-augmented generation for knowledge-intensive nlp tasks,'' \emph{Advances in Neural Information Processing Systems}, vol.~33, pp. 9459--9474, 2020.

\bibitem{houlsby2019parameter}
N.~Houlsby, A.~Giurgiu, S.~Jastrzebski, B.~Morrone, Q.~De~Laroussilhe, A.~Gesmundo, M.~Attariyan, and S.~Gelly, ``Parameter-efficient transfer learning for nlp,'' in \emph{International conference on machine learning}.\hskip 1em plus 0.5em minus 0.4em\relax PMLR, 2019, pp. 2790--2799.

\bibitem{lester2021power}
B.~Lester, R.~Al-Rfou, and N.~Constant, ``The power of scale for parameter-efficient prompt tuning,'' \emph{arXiv preprint arXiv:2104.08691}, 2021.

\bibitem{valipour2022dylora}
M.~Valipour, M.~Rezagholizadeh, I.~Kobyzev, and A.~Ghodsi, ``Dylora: Parameter efficient tuning of pre-trained models using dynamic search-free low-rank adaptation,'' \emph{arXiv preprint arXiv:2210.07558}, 2022.

\bibitem{dettmers2024qlora}
T.~Dettmers, A.~Pagnoni, A.~Holtzman, and L.~Zettlemoyer, ``Qlora: Efficient finetuning of quantized llms,'' \emph{Advances in Neural Information Processing Systems}, vol.~36, 2024.

\bibitem{goo2018abstractive}
C.-W. Goo and Y.-N. Chen, ``Abstractive dialogue summarization with sentence-gated modeling optimized by dialogue acts,'' in \emph{2018 IEEE Spoken Language Technology Workshop (SLT)}.\hskip 1em plus 0.5em minus 0.4em\relax IEEE, 2018, pp. 735--742.

\bibitem{li2019keep}
M.~Li, L.~Zhang, H.~Ji, and R.~J. Radke, ``Keep meeting summaries on topic: Abstractive multi-modal meeting summarization,'' in \emph{Proceedings of the 57th Annual Meeting of the Association for Computational Linguistics}, 2019, pp. 2190--2196.

\bibitem{zhang2018learning}
Y.~Zhang, D.~Y. Ding, T.~Qian, C.~D. Manning, and C.~P. Langlotz, ``Learning to summarize radiology findings,'' \emph{arXiv preprint arXiv:1809.04698}, 2018.

\bibitem{finley2018automated}
G.~Finley, E.~Edwards, A.~Robinson, M.~Brenndoerfer, N.~Sadoughi, J.~Fone, N.~Axtmann, M.~Miller, and D.~Suendermann-Oeft, ``An automated medical scribe for documenting clinical encounters,'' in \emph{Proceedings of the 2018 Conference of the North American Chapter of the Association for Computational Linguistics: Demonstrations}, 2018, pp. 11--15.

\bibitem{enarvi2020generating}
S.~Enarvi, M.~Amoia, M.~D.-A. Teba, B.~Delaney, F.~Diehl, S.~Hahn, K.~Harris, L.~McGrath, Y.~Pan, J.~Pinto \emph{et~al.}, ``Generating medical reports from patient-doctor conversations using sequence-to-sequence models,'' in \emph{Proceedings of the first workshop on natural language processing for medical conversations}, 2020, pp. 22--30.

\bibitem{krishna2020generating}
K.~Krishna, S.~Khosla, J.~P. Bigham, and Z.~C. Lipton, ``Generating soap notes from doctor-patient conversations using modular summarization techniques,'' \emph{arXiv preprint arXiv:2005.01795}, 2020.

\bibitem{ramprasad2023generating}
S.~Ramprasad, E.~Ferracane, and S.~P. Selvaraj, ``Generating more faithful and consistent soap notes using attribute-specific parameters,'' in \emph{Machine Learning for Healthcare Conference}.\hskip 1em plus 0.5em minus 0.4em\relax PMLR, 2023, pp. 631--649.

\bibitem{radford2019language}
A.~Radford, J.~Wu, R.~Child, D.~Luan, D.~Amodei, I.~Sutskever \emph{et~al.}, ``Language models are unsupervised multitask learners,'' \emph{OpenAI blog}, vol.~1, no.~8, p.~9, 2019.

\bibitem{robertson1995okapi}
S.~E. Robertson, S.~Walker, S.~Jones, M.~M. Hancock-Beaulieu, M.~Gatford \emph{et~al.}, ``Okapi at trec-3,'' \emph{Nist Special Publication Sp}, vol. 109, p. 109, 1995.

\bibitem{karpukhin2020dense}
V.~Karpukhin, B.~O{\u{g}}uz, S.~Min, P.~Lewis, L.~Wu, S.~Edunov, D.~Chen, and W.-t. Yih, ``Dense passage retrieval for open-domain question answering,'' \emph{arXiv preprint arXiv:2004.04906}, 2020.

\bibitem{tripathi2023experimental}
H.~Tripathi, ``Experimental approach toward training and analysing siamese deep neural network for sentence with no repeated expressions,'' in \emph{2023 14th International Conference on Computing Communication and Networking Technologies (ICCCNT)}.\hskip 1em plus 0.5em minus 0.4em\relax IEEE, 2023, pp. 1--5.

\bibitem{cormack2009reciprocal}
G.~V. Cormack, C.~L. Clarke, and S.~Buettcher, ``Reciprocal rank fusion outperforms condorcet and individual rank learning methods,'' in \emph{Proceedings of the 32nd international ACM SIGIR conference on Research and development in information retrieval}, 2009, pp. 758--759.

\bibitem{hu2021lora}
E.~J. Hu, Y.~Shen, P.~Wallis, Z.~Allen-Zhu, Y.~Li, S.~Wang, L.~Wang, and W.~Chen, ``Lora: Low-rank adaptation of large language models,'' \emph{arXiv preprint arXiv:2106.09685}, 2021.

\bibitem{meta}
Meta, ``How to fine-tune: Focus on effective datasets,'' \url{https://ai.meta.com/blog/how-to-fine-tune-llms-peft-dataset-curation/}, 2024.

\bibitem{alpaca}
R.~Taori, I.~Gulrajani, T.~Zhang, Y.~Dubois, X.~Li, C.~Guestrin, P.~Liang, and T.~B. Hashimoto, ``Stanford alpaca: An instruction-following llama model,'' \url{https://github.com/tatsu-lab/stanford_alpaca}, 2023.

\bibitem{lin2004rouge}
C.-Y. Lin, ``Rouge: A package for automatic evaluation of summaries,'' in \emph{Text summarization branches out}, 2004, pp. 74--81.

\bibitem{zhang2019bertscore}
T.~Zhang, V.~Kishore, F.~Wu, K.~Q. Weinberger, and Y.~Artzi, ``Bertscore: Evaluating text generation with bert,'' \emph{arXiv preprint arXiv:1904.09675}, 2019.

\bibitem{deutsch2022re}
D.~Deutsch, R.~Dror, and D.~Roth, ``Re-examining system-level correlations of automatic summarization evaluation metrics,'' \emph{arXiv preprint arXiv:2204.10216}, 2022.

\bibitem{savkov2022consultation}
A.~Savkov, F.~Moramarco, A.~P. Korfiatis, M.~Perera, A.~Belz, and E.~Reiter, ``Consultation checklists: Standardising the human evaluation of medical note generation,'' \emph{arXiv preprint arXiv:2211.09455}, 2022.

\bibitem{mchugh2012interrater}
M.~L. McHugh, ``Interrater reliability: the kappa statistic,'' \emph{Biochemia medica}, vol.~22, no.~3, pp. 276--282, 2012.

\end{thebibliography}



\end{document}